\documentclass{amia}
\usepackage{appendix}
\usepackage[labelfont=bf]{caption}
\usepackage[superscript,nomove]{cite}
\usepackage{color}
\usepackage[ruled,linesnumbered]{algorithm2e}

\usepackage{import}
\usepackage[english]{babel}
\usepackage{graphicx}
\usepackage{booktabs}
\usepackage{multirow}
\usepackage[table,xcdraw]{xcolor}
\usepackage{epstopdf}
\usepackage{pdfpages}
\usepackage{amsmath,amssymb,multicol}

\usepackage{mathtools}
\usepackage{xr-hyper}
\usepackage{bigstrut}
\usepackage{multirow}
\usepackage{bm}
\usepackage[utf8]{inputenc}
\usepackage{tikz}
\usetikzlibrary{matrix,shapes,arrows,positioning,chains}
\usepackage{adjustbox}
\usepackage{url}
\usepackage{subfig}
\usepackage{caption}
\usepackage{wrapfig}
\usepackage{array}
\newtheorem{theorem}{Theorem}
\usepackage[skip=4pt]{caption}
\usepackage{etoolbox}
\usepackage{url}
\usepackage{helvet}
\usepackage[T1]{fontenc}
\usepackage[utf8]{inputenc}
\DeclareUnicodeCharacter{00A0}{ }

\makeatletter
\def\thickhline{%
  \noalign{\ifnum0=`}\fi\hrule \@height \thickarrayrulewidth \futurelet
   \reserved@a\@xthickhline}
\def\@xthickhline{\ifx\reserved@a\thickhline
               \vskip\doublerulesep
               \vskip-\thickarrayrulewidth
             \fi
      \ifnum0=`{\fi}}
\makeatother

\newlength{\thickarrayrulewidth}
\begin{document}

\title{MD-MTL: An Ensemble Med-Multi-Task Learning Package for Disease Scores Prediction and Multi-Level Risk Factor Analysis}

\author{Lu Wang, PhD$^{1}$, Haoyan Jiang, BS$^{1}$, Mark Chignell, PhD$^{1}$}

\institutes{
    $^1$Dept. of Mechanical and Industrial Engineering, University of Toronto, Toronto, ON, Canada \\
}

\maketitle

\noindent{\bf Abstract}

\textit{While many machine learning methods have been used for medical prediction and risk factor analysis on healthcare data, most prior research has involved single-task learning (STL) methods. However, healthcare research often involves multiple related tasks. For instance, implementation of disease scores prediction and risk factor analysis in multiple subgroups of patients simultaneously and risk factor analysis at multi-levels synchronously. In this paper, we developed a new ensemble machine learning Python package based on multi-task learning (MTL), referred to as the Med-Multi-Task Learning (MD-MTL) package and applied it in predicting disease scores of patients, and in carrying out risk factor analysis on multiple subgroups of patients simultaneously. Our experimental results on two datasets demonstrate the utility of the MD-MTL package, and show the advantage of MTL (vs. STL), when analyzing data that is organized into different categories (tasks, which can be various age groups, different levels of disease severity, etc.). 
}

\section*{Introduction}
\begin{figure} 
    \centering
\subfloat[STL Model]{%
        \includegraphics[width=0.5\columnwidth]{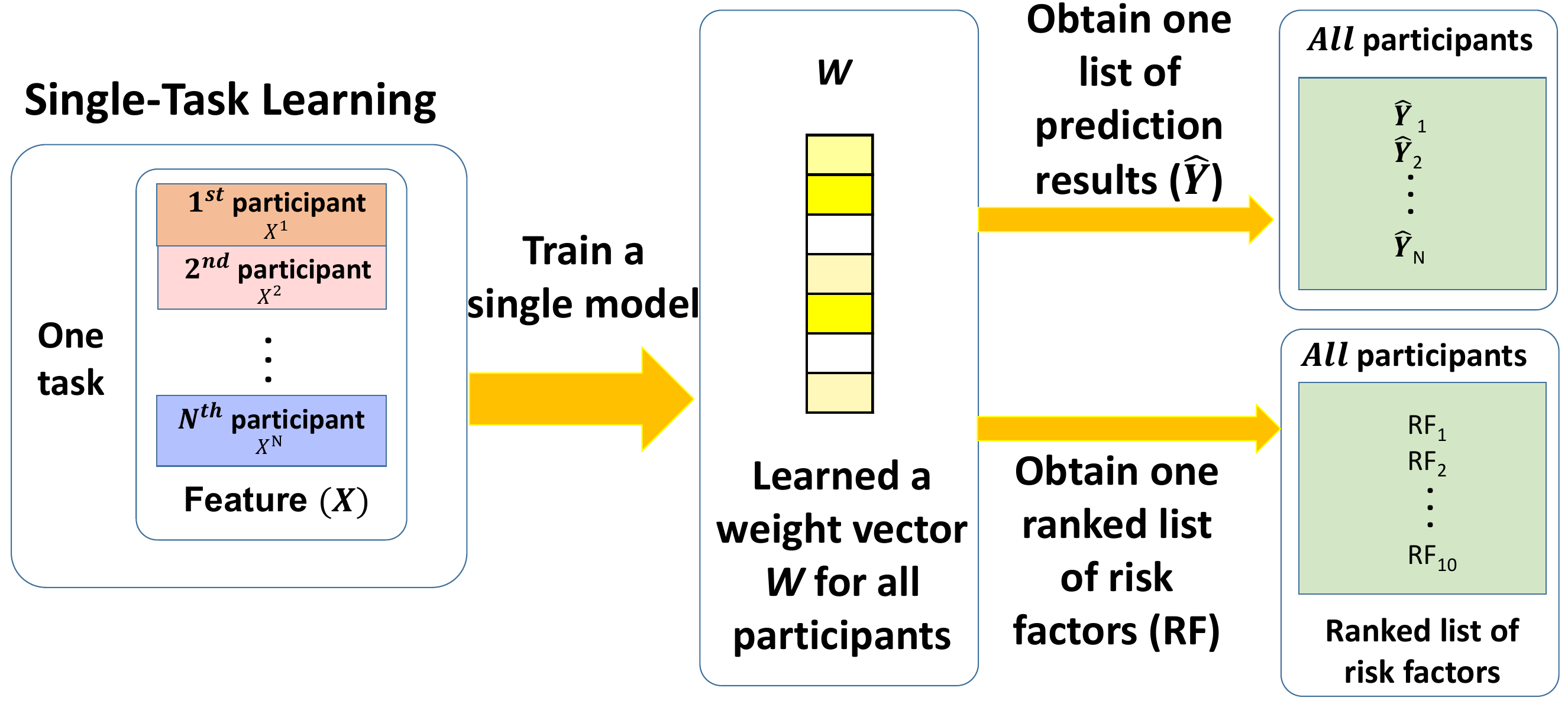}}
         \hfill
  \subfloat[MTL Model]{%
        \includegraphics[width=0.5\columnwidth]{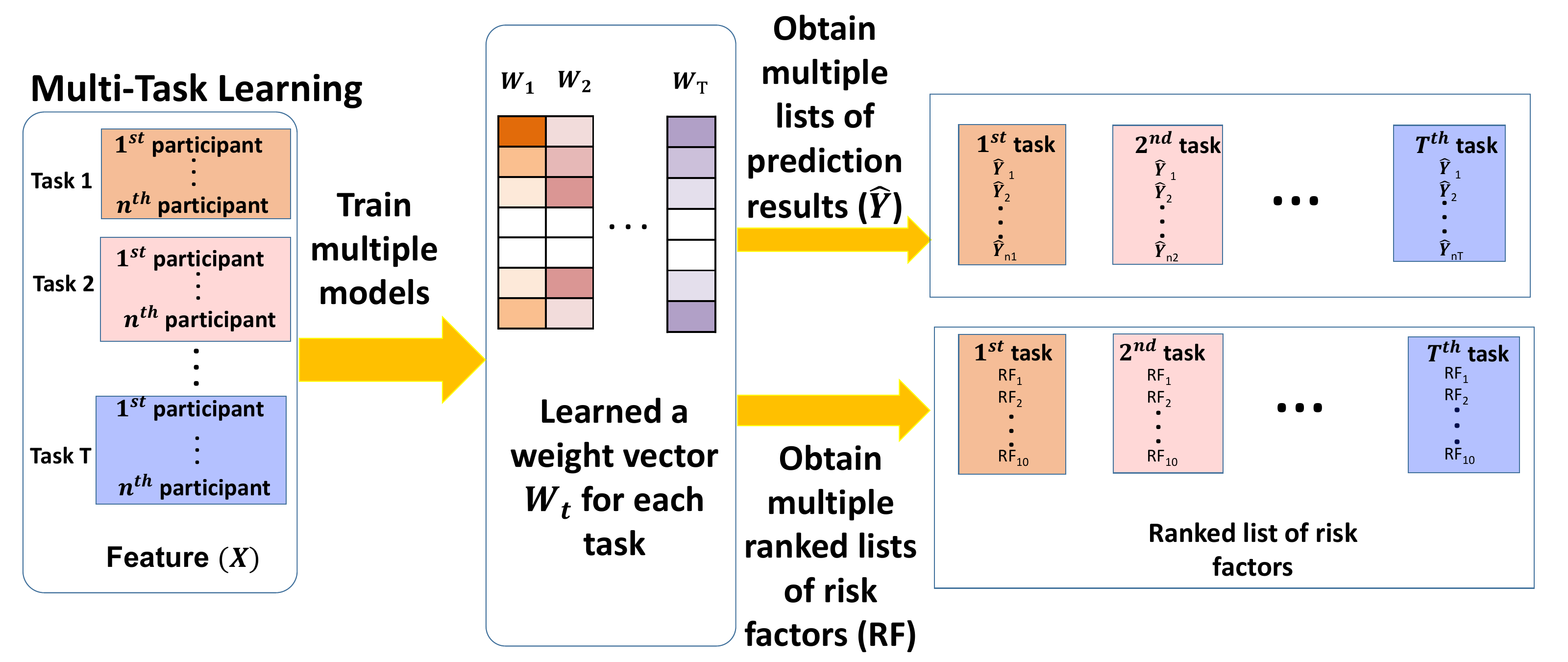}}
    \hfill
    \subfloat[CMTL Model]{%
        \includegraphics[width=0.7\columnwidth]{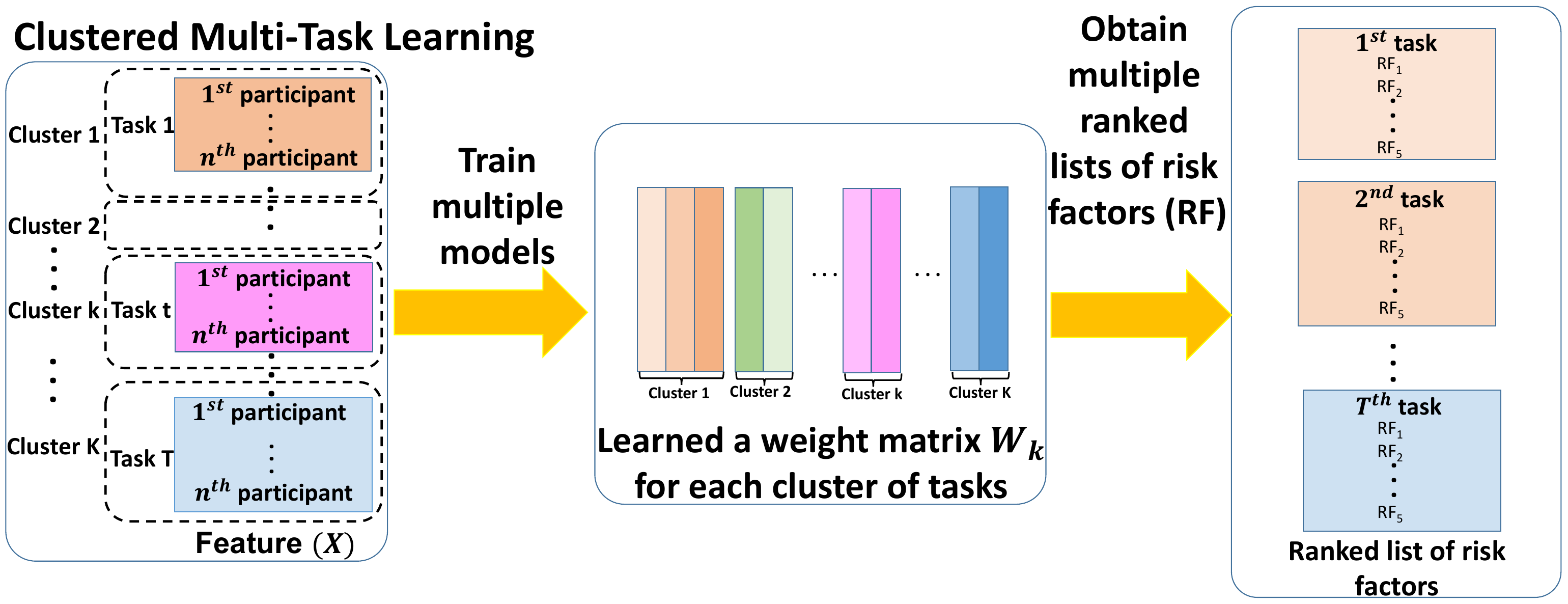}}
    
  \caption{\label{fig:idea}Prediction of disease scores and multi-level risk factor analysis are implemented for health data within the MTL framework: 1) MTL model trains multiple models simultaneously and learns a ranked list of RFs for each subpopulation. 2) CMTL model clusters subpopulations into several groups and obtains multiple ranked lists of RFs for all subpopulations. But STL family models merely train a global model that is one-size-fits-all at the population-level only. (Darker box in the weight vector/matrix means higher value of feature weight.)
  }
  \label{fig1} 
\end{figure}

In recent years there has been a sharp increase in the volume of health data involving meaningful categories of patients, e.g., disease scores prediction \cite{ref1,ref2,ref3,ref4} and risk factor analysis \cite{ref5,ref6,ref7,ref8} in multiple subgroups of patients. Traditional machine learning methods either build a global model at the population-level only or build a local model for each subpopulation. These are single-task learning (STL) approaches (Figure \ref{fig:idea}(a)). These STL approaches have been shown effective initial efforts in disease scores prediction and risk factor analysis. However, they have the following disadvantages: 1) A global model fails to capture the data heterogeneity in the population. 2) A local model fails to utilize the shared information among subpopulations. In addition, an over-parameterized model is susceptible to overfitting especially when the sample size of a subpopulation is small.

STL trains a model for a subpopulation independently, ignoring relations that are shared between subpopulations, leading to over-fitting of the model \cite{ref13}. Alternatively, STL may train a model for pooled subpopulations, ignoring the heterogeneous property of data \cite{ref13}. 
In contrast, multi-task learning (MTL) provides methods to train a model for each subpopulation utilizing appropriate shared information across tasks. MTL is capable of training multiple related models for all subpopulations at the same time by utilizing shared information among these subpopulations. Thus, MTL can predict the disease scores and learn multiple ranked lists of risk factors (RFs) in multiple subgroups of patients simultaneously. To take into account both data heterogeneity and homogeneity, MTL model is implemented to build multiple related models along with an across-all-tasks penalty/regularization term, i.e., $l_{2,1}-$norm, to ensure that the weight of each input feature is either small or large for all subpopulations \cite{ref9}, so that the ranked list of RFs at population-level can be learned (Figure \ref{fig:idea}(b)).

In the real-world scenario, grouping structure often exists in multiple related tasks. The clustered multi-task learning (CMTL) model is used to reveal the grouping structure of tasks and learn multiple related tasks simultaneously \cite{ref10}, combining clustering, prediction and feature selection within the MTL framework (Figure \ref{fig:idea}(c)). 

Using the MTL and CMTL models, prediction of disease scores in multiple subgroups of patients \cite{ref11} and multi-level risk factor analysis \cite{ref12} (e.g., three levels in our paper: each subpopulation, each subgroup of the population and the whole population) were carried out in the research reported below. The main contributions of this paper are as follows:
\begin{itemize}
\itemsep0em
\item Development of an open-source ensemble machine learning Python package conducting multi-med tasks for health data using MTL family models, i.e., Med-Multi-Task Learning (MD-MTL)\footnote{\url{https://github.com/Interactive-Media-Lab-Data-Science-Team/Vampyr-MTL}}.
\item Experimental analysis of a semi-public medical dataset (i.e., needing approval from the data hosting organization and a research ethics board) and a  public behavioral dataset with unbalanced subgroup sizes.
\item Disease score prediction and risk factor analysis in multiple subgroups of patients simultaneously using the MTL model.
\item Identification of population-level RFs and ranked lists of RFs for each subpopulation using the CMTL model.
\end{itemize} 

\vspace{-0.5cm}
\section*{Methods}
\label{method}
We developed a Python package using multi-task learning (MTL), called MD-MTL, to implement health data analysis in terms of disease scores prediction and medical risk factor analysis. We compared the quality of prediction for MTL and STL models with respect to a key outcome in each of two datasets, using mean absolute error (MAE) as the evaluative criterion. We also used the CMTL model to identify risk factors at different levels of the hierarchically organized data. 
\subsection*{Data Description and Task Definition}
The Framingham Heart Study (\textbf{FHS}) is a relatively small semi-public medical benchmark dataset that requires organizational approval for us to use, while the Behavioral Risk Factor Surveillance System (\textbf{BRFSS}) is a large public dataset. We used two datasets to predict two aspects of psychological status, predicting cognitive scores, and the number of days where self-reported mental health was not good, in the FHS and BRFSS datasets, respectively. We implemented both prediction of cognitive scores and \textbf{\textit{two-level risk factor analysis (i.e., subpopulation and population-levels)}} using the MTL model for the FHS dataset. We conducted both prediction of the number of days with not good mental health using the MTL model and \textbf{\textit{three-level risk factor analysis (i.e., subpopulation, population and cluster-of-subpopulation-levels)}} using the CMTL model for the BRFSS dataset. 

We divided these two datasets into training and testing sets using stratified sampling with shuffling. $60\%$ of instances were used for training and the rest for testing. All the data were preprocessed with MinMax scaling between 0 and 1.
\vspace{-0.5cm}
\paragraph*{Framingham Heart Study (FHS)}
The original goal of FHS was to identify the shared risk factors contributing to cardiovascular disease. Launched in 1948, FHS is a multigenerational study of family patterns of cardiovascular and other diseases. We studied the effect of heart disease on cognitive health, and the relationship between heart disease and cognitive health, by predicting the \textbf{\textit{COGSCR}} (i.e., cognitive function test) score as the outcome, using predictor variables relating to heart disease related and physical health tests. For these analyses we merged the Initial Congestive Heart Failure (CHF) Cases Reviewed for Additional Information file with the Cognitive Function Test - Mini Mental State Exam (MMSE).

The dataset originally contained $3,552$ instances and $114$ variables. We focused on the total cognitive score and removed observations with missing outcome values as well as index variables that did not have a potential predictive relationship with the outcome. The resulting dataset used in our analysis included 633 instances and 79 predictor variables relating to physical health and measurement of CHF, e.g., blood pressure, blood glucose levels. 

Using the multi-task approach, we divided the patients into four subgroups (tasks) based on the Valvular Heart Disease (\textbf{\textit{VHD}}) level. Hence, the tasks were defined in FHS based on different VHD level as shown in the first column in Table \ref{fhs1}, i.e., VHD\_ 0, VHD\_1, VHD\_ 2 and VHD\_ 3. In addition to predicting total cognitive scores, we also conducted two-levels (i.e., at subpopulation and population-levels) risk factor analysis.

\vspace{-0.5cm}
\paragraph*{Behavioral Risk Factor Surveillance System (BRFSS)}
The BRFSS dataset is a collaborative project between all the states in the U.S. and the Centers for Disease Control and Prevention (CDC), and aims to collect uniform, state-specific data on preventable health practices and risk behaviors that affect the health of the adult population (i.e., adults aged 18 years and older). In the experiment, we used the BRFSS dataset that was collected in 2018\footnote{\url{https://www.cdc.gov/brfss/annual_data/annual_2018.html}}. The BRFSS dataset was collected via the phone-based surveys with adults residing in private residence or college housing.


Our analysis predicted \textbf{\textit{MENTHLTH}}, i.e., the number of days that self-report mental health was not good during the past 30 days. The original BRFSS dataset contained 437, 436 instances and 275 variables. After deleting entries with missing outcome values, the preprocessed dataset contained 119, 929 observations with 90 predictor variables and one output variable, i.e., \textbf{\textit{MENTHLTH}}. We examined \textbf{\textit{two settings of task definition}}: 1). For \textbf{\textit{cognitive scores prediction and two-levels risk factor analysis}}, we defined six tasks based on the age groups, i.e.,18 years to 24 years, 25 years to 34 years, 35 years to 44 years, 45 years to 54 years, 55 years to 64 years and equal or older than 65 years shown as the first column in Table \ref{brfss1}.
2). For \textbf{\textit{three-level risk factor analysis}}, at the most detailed level the tasks were defined in BRFSS based on the geographic information, i.e, 50 states in the U.S. plus the Washington, District of Columbia (DC) district in the U.S., so that there were 51 related tasks, i.e., 51 subpopulations. At a higher level, the tests were defined in terms of four regions that the U.S. states were grouped into based on cluster analysis of the data (Figure \ref{fig:r4}). 
\vspace{-0.5cm}
\subsection*{MTL framework}
Many independent tasks are rare compared with multiple related ones in most real-world applications, so that MTL was implemented to capture the information in the multiple related tasks. We assumed that all the tasks were related since they shared the same feature space. To encode the task relatedness into the objective function of MTL, a penalty/regularization term across all the tasks, denoted as $\Omega (\Phi)$, is used. Therefore, minimization of penalized empirical loss is expressed as the framework's objective formulation:
\begin{equation}
\label{eq1}
\min_{\Phi}\mathcal{L}(\Phi)+\Omega (\Phi),
\end{equation}
where the empirical loss function is denoted as $\mathcal{L}(\Phi)$.
\vspace{-0.5cm}
\paragraph*{Algorithm of disease scores prediction and risk factor analysis using MTL model}
The loss function in MTL is formulated as:
\begin{equation}
\label{eq2}
\mathcal{L}(\Phi)=\frac{1}{2}\sum_{t=1}^T \left \|X_t \Phi_t^T -Y_t  \right \|^2,
\end{equation}
where $T$ is the number of tasks and its corresponding index number is $t$. $\Phi\in R^{T\times J}$ is the weight matrix of $J$ continuous input variables. $X$ is the input matrix and the $t^{th}$ task has the input matrix denoted as $X_t \in \mathbb{R}^{n_t \times J}$. $Y_t$ denotes the output variable.

As we mentioned in Section Introduction, $l_{2,1}-$norm is the penalty/regularization term across all the tasks to encode the joint sparsity:
\begin{align}
\label{eq3}
&\Omega (\Phi)=\lambda \left \|\Phi \right \|_{2,1}=\lambda\sum_{j=1}^{J}\sqrt{\sum_{t=1}^{T}\left | \phi_{tj} \right |^2},
\end{align}
where $j$ is the corresponding index number of continuous input variables and $\phi_{tj}$ denotes weight scalar of the $t^{th}$ task's $j^{th}$ feature. 

As a result, the objective function of MTL can be re-written as:
\begin{equation}
\label{eq4}
\min_{\Phi}\frac{1}{2}\sum_{t=1}^T \left \|X_t\Phi_t^T -Y_t  \right \|^2+\lambda \sum_{j=1}^{J}\sqrt{\sum_{t=1}^{T}\left | \phi_{tj} \right |^2},
\end{equation}
where $\lambda \geq 0$, called tuning parameter, can be used to adjust the penalty/regularization term and control the sparsity of feature weights matrix. It produces more sparse feature weights matrix when the value of $\lambda$ is increasing. 

\vspace{-0.5cm}
\paragraph*{Algorithm of multi-level risk factor analysis using CMTL model}
Multiple tasks in the real-world applications not only are related, but also show a more complicated grouping structure, which can be seen from that the estimated weights of tasks from the same group are closer than these from distinct groups. To reveal the grouping structure of multiple tasks, $K$-means clustering is employed by implementing CMTL. To encode the grouping structure of multiple tasks in the formulation, $K$-means's sum-of-squares error (SSE) is used as the regularization term in the objective function of CMTL.

We assume that $T$ tasks can be clustered into $K$ clusters, where $K<T$. The cluster's corresponding index number is $k$ and the index set is defined as $\mathcal{I}_k=\{1,2,\cdots,K\}$. Let $\bar{\Phi}_k=\frac{1}{n_k}\sum_{k \in \mathcal{I}_k}\Phi_k$ be the mean function of the weight vectors in the $k^{th}$ cluster, so that the SSE is calculated as: 
\begin{align}
\label{eq:sse}
\sum_{k=1}^K\sum_{k \in \mathcal{I}_k}\parallel \Phi_k- \bar{\Phi}_k \parallel_2^2=\text{tr}(\Phi\Phi^T)-\text{tr}(\Phi OO^T\Phi^T),
\end{align}
where tr$(\cdot)$ is the trace norm of matrix and $O\in \mathbb{R}^{T \times K}$ is the cluster indicator matrix that is orthogonal:
\begin{equation}
O_{t,k}=\left\{ \begin{array}{rcl}
\frac{1}{\sqrt{n_k}} &\mbox{if}& t \in \mathcal{I}_k, \\
0 &\mbox{if}& t \notin \mathcal{I}_k,
\end{array}\right.
\end{equation}
where $n_k$ is the number of input instances/participants in cluster $k$.
Since the orthogonal cluster indicator matrix $O$ is non-convex that exhibits the above mentioned special structure, the SSE in Eq.(\ref{eq:sse}) is hard to be minimized. To overcome this issue, we use a spectral relaxation approach, the latter is expressed as $O^TO=I_K$. Furthermore, a convex relaxation that relaxes the feasible domain of $OO^T$ into a convex set is proposed in \cite{c15}, i.e., $\mathcal{C}=\{C| \text{tr}(C)=T, C \preceq I, C\in \mathbb{S}^T_{+} \}$, where $\mathbb{S}^T_{+}$ is a subset of positive-semidefinite matrices. As a result, $OO^T$ can be approximated through the convex set $C$. In conclusion, the previously mentioned two types of relaxation methods generate the convex relaxed CMTL expressed as:
\begin{align}
\label{eq:cCMTL}
& \underset{\Phi,C}{\text{min}} &\!\!\!\!\!\! &\mathcal{L}(\Phi)+\rho_1[\text{tr}(\Phi\Phi^T)-\text{tr}(\Phi C\Phi^T)]+\rho_2 \text{tr}(\Phi\Phi^T),\nonumber\\
& \text{s. t.} &\!\!\!\! \!\! &  \text{tr}(C)=K, \Phi \preceq I, \Phi\in \mathbb{S}^T_{+}
\end{align}
where $\text{tr}(\Phi\Phi^T)=|| \Phi || _O ^2$  is used to shrink the weights and relieve the multicollinearity, which is also called the square of Frobenius norm of $\Phi$. A parameter $\eta$ is introduced, which is defined as $\eta=\frac{\rho_2}{\rho_1}>0$. Then with some simple algebraic calculations, the convex relaxed CMTL is reformulated as:
\begin{align}
\label{eq:cCMTL1}
& \underset{\Phi,C}{\text{min}} &\!\!\!\!\!\! &\sum_{t=1}^T\!\! \frac{1}{N_t} \sum_{i=1}^{N_t}(X_i^t (C_i^t)^T -Y_i^t )^2 
+ \rho_1\eta(1+\eta)\text{tr}(\Phi(\eta I+C)^{-1}\Phi^T),\nonumber\\
& \text{s. t.} &\!\!\!\! \!\! &  \text{tr}(C)=K, C \preceq I, C\in \mathbb{S}^T_{+}
\end{align}
where $N_t$ is the number of instances/participants in task $t$ and $i$ is the index of instance/participant in the $t^{th}$ task. Pease refer to the online Appendices\footnote{\url{https://github.com/wanglu61/MD-MTL-supportive-materials}} for the optimization of algorithms.
\vspace{-0.5cm}

\subsection*{Models Implementation}
We used the MD-MTL package that we developed  to implement the MTL and CMTL models, calling the MD-MTL functions \textit{functions.MTL\_Least\_L21()} and \textit{functions.MTL\_Cluster\_Least\_L21()} after importing the module by \textit{from MD-MTL import functions}. 
In addition to MTL, STL methods were applied under two settings: 1) Global setting, i.e., a prediction model was trained for all tasks; 2) Individual setting, i.e., a prediction model was trained for each task separately. In the individual setting the heterogeneity among tasks was fully considered but not the task relatedness; in the global setting all the heterogeneities were neglected. 

We selected seven commonly used STL based regression methods mentioned in Section Results, which were implemented in Python using the package \textit{scikit-learn} \cite{sklearn}: 1) Linear regression (Linear) was trained using the \textit{linear\_model.LinearRegression()} function to minimize the residual sum of squares between the observed targets in the dataset, and the targets predicted by the linear approximation. 2) Ridge regression (Ridge) was trained using \textit{linear\_model.Ridge()} function to minimize a penalized residual sum of squares. 3). Multi-layer perceptron (MLP) was trained using \textit{sklearn.neural\_network.MLPRegressor()} to optimize the squared-loss.
4). Lasso regression (Lasso) was trained using \textit{linear\_model.Lasso()} function to estimate sparse coefficients.
5). LassoLars was trained using \textit{linear\_model.LassoLars()} function, which is piecewise linear as a function of the norm of its coefficients.
6). Bayesian Ridge regression (BayesianRidge) was trained using \textit{linear\_model.BayesianRidge()} function, which estimates a probabilistic model of the regression problem.
7). Linear regression with Elastic-Net (ElasticNet) was trained using \textit{linear\_model.ElasticNet()} that is trained with both $l_1$ and $l_2$-norm regularization of the coefficients.

\section*{Results}
\label{results}

\paragraph*{Prediction Performance Comparisons Results}
To evaluate the overall performance of each method, we used mean absolute error (MAE), where MAE is defined as: a set of forecasts, which is negatively-oriented score (i.e., lower is better) and ranges from 0 to $\infty$. To formally define MAE, we use $y_i$ and $\hat{y}$ to represent the predicted and the original values, respectively. MAE=$\frac{1}{N} \sum_{i=1}^{N} \left |y_i-\hat{y}  \right |$, where $i$ is the index of each patient and N is the number of patients.  

Prediction results in terms of MAE for the different models for the two datasets, \textbf{\textit{FHS}} and \textbf{\textit{BRFSS}}, are shown in Table \ref{fhs1} and Table \ref{brfss1}, respectively. 
\begin{table}[ht!]
\centering
 \caption{The mean absolute error (MAE) of our proposed multi-task learning (MTL) model and seven single-task learning (STL) models in both global and individual settings for dataset Framingham Heart Study (\textbf{FHS}). Please refer to the last subsection of Section Methods, i.e., Models Implementation, for details about the seven STL models and definition of global and individual settings. The \textbf{\textit{first and second columns}} represent the heart disease group (HDG) of each task and number of patients in each task, respectively. The \textbf{\textit{third column}} contains both mean and standard deviation (SD) of the outcome variable (i.e., total score for cognitive function test of each patient) values as the primary statistical information in each task. Total indicates the whole population instead of a single task name. Please refer to the first subsection of Section Methods, i.e., Data Description and Task Definition, for details about the task definition and outcome. Note that, \textbf{\textit{SD}} is shown in parentheses at the second row in each cell that is under the MAE. The best performance results for each task/row are shown in \textbf{boldface}.\vspace{0.2cm}}
\label{fhs1}
\resizebox{\textwidth}{!}{%
\begin{tabular}{|l|l|l|l|lllllll|lllllll|l}
\cline{1-18}
  &&&& \multicolumn{7}{c|}{Global Setting} &\multicolumn{7}{c|}{Individual Setting} \\ \cline{5-18} 
\multirow{-2}{*}{\begin{tabular}[c]{@{}l@{}}Task\\ (HDG)\end{tabular}} &\multirow{-2}{*}{\begin{tabular}[c]{@{}l@{}}No. of \\patients\end{tabular}} &\multirow{-2}{*}{\begin{tabular}[c]{@{}l@{}}Mean\\ /SD\end{tabular}} & \multirow{-2}{*}{MTL} &
  Linear &Ridge &MLP &Lasso &\begin{tabular}[c]{@{}l@{}}Lasso\\ Lars\end{tabular} &\begin{tabular}[c]{@{}l@{}}Bayesian\\ Ridge\end{tabular} &\begin{tabular}[c]{@{}l@{}}Elastic\\ Net\end{tabular} &Linear &Ridge &MLP &Lasso &\begin{tabular}[c]{@{}l@{}}Lasso\\ Lars\end{tabular} &\begin{tabular}[c]{@{}l@{}}Bayesian\\ Ridge\end{tabular} &\begin{tabular}[c]{@{}l@{}}Elastic\\ Net\end{tabular} &\\ \cline{1-18}
&& 28.4 & \textbf{1.38} & 5.67 & 11.24 & 3.82 & 1.50 & 1.50 & 1.50 & 1.50 & 1.48E+12 & 16.13 & 3.56 & 1.46 & 1.46 & 1.46 & 1.46 & \\
\multirow{-2}{*}{VHD\_0} & \multirow{-2}{*}{220} &
  (1.95) & { (0.090)} & { (2.86)} & { (0.29)} & { (0.34)} & { (0.13)} & { (0.13)} & { (0.13)} & { (0.13)} & { (6.58E+11)} & { (0.27)} & { (0.33)} & { (0.12)} & { (0.12)} & { (0.12)} & { (0.12)} & \\ \cline{1-18}
&&28.72 & \textbf{0.92} & 19.63 & 12.77 & 4.22 & 1.39 & 1.39 &1.39 &1.39 &5.49E+11 &23.86 & 2.72 & 1.21 & 1.21 & 1.21 & 1.21 & \\
\multirow{-2}{*}{VHD\_1} & \multirow{-2}{*}{61} &
  { (1.43)} & { (0.14)} & { (3.86)} & { (0.4)} & { (0.41)} &  { (0.15)} & { (0.15)} & { (0.15)} & { (0.15)} &{ (8.13E+11)} &{ (0.31)} &{ (0.44)} &{ (0.13)} &{ (0.13)} &{ (0.13)} &{ (0.13)} &{ } \\ \cline{1-18}
&&28.03 &1.47 &3.49 &15.87 &6.90 &\textbf{1.38} &\textbf{1.38} &\textbf{1.38} &\textbf{1.38} &1.46 &20.75 &3.34 &1.40 &1.40 &1.40 &1.40 &\\
\multirow{-2}{*}{VHD\_2} &\multirow{-2}{*}{129} &
  { (1.7)} &{ (0.092)} &{ (2.71)} &{ (0.22)} &{ (0.24)} &{ (0.14)} &{ (0.14)} &{ (0.14)} &{ (0.14)} &{ (0.09)} &{ (0.26)} &{ (0.29)} &{ (0.14)} &{ (0.14)} &{ (0.14)} &{ (0.14)} &{ } \\ \cline{1-18}
&&27.96 &\textbf{1.54} &3.98 &11.14 &3.26 &1.63 &1.63 &1.63 &1.63 &1.22E+12 &15.93 &2.87 &1.66 &1.66 &1.66 &1.66 &\\
\multirow{-2}{*}{VHD\_3} &\multirow{-2}{*}{223} &
  { (2.23)} &{ (0.16)} &{ (2.34)} &{ (0.13)} &{ (0.34)} &{ (0.14)} &{ (0.14)} &{ (0.14)} &{ (0.14)} &{ (1.43E+12)} &{ (0.15)} &{ (0.26)} &{ (0.11)} &{ (0.11)} &{ (0.11)} &{ (0.11)} &\\ \cline{1-18}
&&28.2 &\textbf{1.41} &1.87 &8.62 &2.53 &1.44 &1.44 &1.44 &1.44 &9.97E+11 &17.76 &3.19 &1.49 &1.49 &1.49 &1.49 &\\
\multirow{-2}{*}{TOTAL} &\multirow{-2}{*}{633} &
  (1.98) &{ (0.063)} &{ (1.15)} &{ (0.58)} &{ (0.16)} &{ (0.07)} &{ (0.07)} &{ (0.07)} &{ (0.07)} &{ (4.72E+11)} &{ (0.1)} &{ (0.16)} &{ (0.09)} &{ (0.09)} &{ (0.09)} &{ (0.09)} &\\ \cline{1-18}
\end{tabular}%
}
\end{table}
\begin{table}[ht!]
\centering
\caption{MAE of the MTL and seven STL models in both global and individual settings for the dataset Behavioral Risk Factor Surveillance System (\textbf{BRFSS}). The \textbf{\textit{first and second columns}} represent the age group (AG) of each task and number of participants in each task of testing dataset, respectively. The \textbf{\textit{third column}} contains both mean and SD of the outcome variable (i.e., number of days mental health that is not good during the past 30 days). (The best performance results are in \textbf{boldface}.)\vspace{0.2cm}}
\label{brfss1}
\resizebox{\textwidth}{!}{%
\begin{tabular}{|l|l|l|l|lllllll|lllllll|l}
\cline{1-18}
 &&&&
 \multicolumn{7}{c|}{Global Setting} &\multicolumn{7}{c|}{Individual Setting} \\ \cline{5-18}  \multirow{-2}{*}{\begin{tabular}[c]{@{}l@{}}Task\\ (AG)\end{tabular}} &\multirow{-2}{*}{\begin{tabular}[c]{@{}l@{}}No. of \\ participants\end{tabular}} &\multirow{-2}{*}{\begin{tabular}[c]{@{}l@{}}Mean\\ /SD\end{tabular}} & \multirow{-2}{*}{MTL} &
  Linear &Ridge &MLP &Lasso &\begin{tabular}[c]{@{}l@{}}Lasso\\ Lars\end{tabular} &\begin{tabular}[c]{@{}l@{}}Bayesian\\ Ridge\end{tabular} &\begin{tabular}[c]{@{}l@{}}Elastic\\ Net\end{tabular} &Linear &Ridge &MLP &Lasso &\begin{tabular}[c]{@{}l@{}}Lasso\\ Lars\end{tabular} &\begin{tabular}[c]{@{}l@{}}Bayesian\\ Ridge\end{tabular} &\begin{tabular}[c]{@{}l@{}}Elastic\\ Net\end{tabular} &\\ \cline{1-18}
&&10.71 &\textbf{3.52} &3.63 &3.91 &3.63 &4.29 &9.04 &3.63 & 6.81 &3.52 &4.5 & \textbf{3.52} &4.17 &8.94 &\textbf{3.52} &6.72 &\\
\multirow{-2}{*}{18-24} &\multirow{-2}{*}{27,858} &
(10.53) &{(0.03)} &{(0.02)} &{(0.02)} &{(0.03)} &{(0.02)} &{(0.02)} &{(0.02)} &{(0.02)} &{(0.03)} &{(0.03)} &{(0.03)} &{(0.02)} &{(0.01)} &{(0.03)} &{(0.02)} &\\ \cline{1-18}
 &&11.95 &\textbf{3.71} &3.74 &3.99 &3.75 &4.46 &9.31 &3.74 &6.97 &3.72 &4.77 &\textbf{3.71} &4.41 & 9.53 &\textbf{3.71} &6.94 &\\
\multirow{-2}{*}{25-34} &\multirow{-2}{*}{24,252} &
{(10.88)} &{(0.01)} &{(0.01)} &{(0.02)} &{(0.02)} &{(0.02)} &{(0.04)} &{(0.01)} &{(0.04)} &{(0.01)} &{(0.02)} &{(0.01)} &{(0.01)} &{(0.03)} &{(0.01)} &{(0.04)} &{} \\ \cline{1-18}
&&11.46 &3.61 &3.61 &3.82 &3.62 &4.34 &9.15 &3.61 &6.83 &\textbf{3.60} &4.76 &\textbf{3.60} &4.30 &9.26 &\textbf{3.60} &6.75 &\\
\multirow{-2}{*}{35-44} &\multirow{-2}{*}{20,570} &
{(10.65)} &{(0.01)} &{(0.02)} &{(0.01)} &{(0.01)} &{(0.01)} &{(0.05)} &{(0.01)} &{(0.04)} &{(0.01)} &{(0.03)} &{(0.01)} &{(0.02)} &{(0.04)} &{(0.01)} &{(0.04)} &{} \\ \cline{1-18}
&&10.53 &3.64 &\textbf{3.59} &3.77 &3.61 &4.04 &8.34 &\textbf{3.59} &6.1 &3.62 &4.68 &3.66 &4.11 &8.17 &3.62 &6.21 &\\
\multirow{-2}{*}{45-54} &\multirow{-2}{*}{17,748} &
(9.80) &{(0.04)} &{(0.05)} &{(0.04)} &{(0.06)} &{(0.05)} &{(0.07)} &{(0.05)} &{(0.06)} &{(0.04)} &{(0.05)} &{(0.06)} &{(0.06)} &{(0.05)} &{(0.04)} &{(0.07)} &\\ \cline{1-18}
&&10.53 &3.51 &\textbf{3.48} &3.70 &3.50 &4.16 & 8.76 & \textbf{3.48} & 6.52 & 3.51 & 4.77 & 3.52 & 4.13 & 8.58 & 3.51 &6.48 & \\
\multirow{-2}{*}{55-64} & \multirow{-2}{*}{17,598} &
(10.18) & {(0.02)} & {(0.02)} & {(0.02)} & {(0.04)} & {(0.01)} & {(0.03)} & {(0.02)} & {(0.03)} & {(0.03)} & {(0.01)} & {(0.03)} & {(0.01)} & {(0.02)} & {(0.02)} & {(0.03)} & \\ \cline{1-18}
&&10.54 & \textbf{3.62} & 3.72 & 3.74 & 3.68 & 3.89 & 8.01 & 3.71 & 5.8 & 4.58 & 4.84 & 4.08 & 4.05 & 7.85 & 3.67 & 6.00 & \\
\multirow{-2}{*}{$\geq$65} & \multirow{-2}{*}{11,903} & 
(9.44) & {(0.07)} & {(0.15)} & {(0.03)} & {(0.08)} & {(0.03)} & {(0.05)} & {(0.14)} & {(0.04)} & {(0.99)} & {(0.17)} & {(0.54)} & {(0.04)} & {(0.04)} & {(0.09)} & {(0.05)} & \\ \cline{1-18}
&&11.02 & \textbf{3.60} & 3.62 & 3.81 & 3.62 & 4.24 & 8.86 & 3.62 & 6.59 & 3.69 & 4.70 & 3.65 & 4.22 & 8.84 & 3.61 & 6.59 & \\
\multirow{-2}{*}{TOTAL} & \multirow{-2}{*}{119,929} &
(10.38) & {(0.01)} & {(0.01)} & {(0.01)} & {(0.01)} & {(0.01)} & {(0.02)} & {(0.01)} & {(0.02)} & {(0.1)} & {(0.02)} & {(0.05)} & {(0.01)} & {(0.01)} & {(0.01)} & {(0.02)} & \\ \cline{1-18}
\end{tabular}%
}
\end{table}

\begin{figure*}[t]
\centering

\subfloat[RFs Representing at Least Two of the Subpopulations]{%
  \includegraphics[width=0.45\columnwidth]{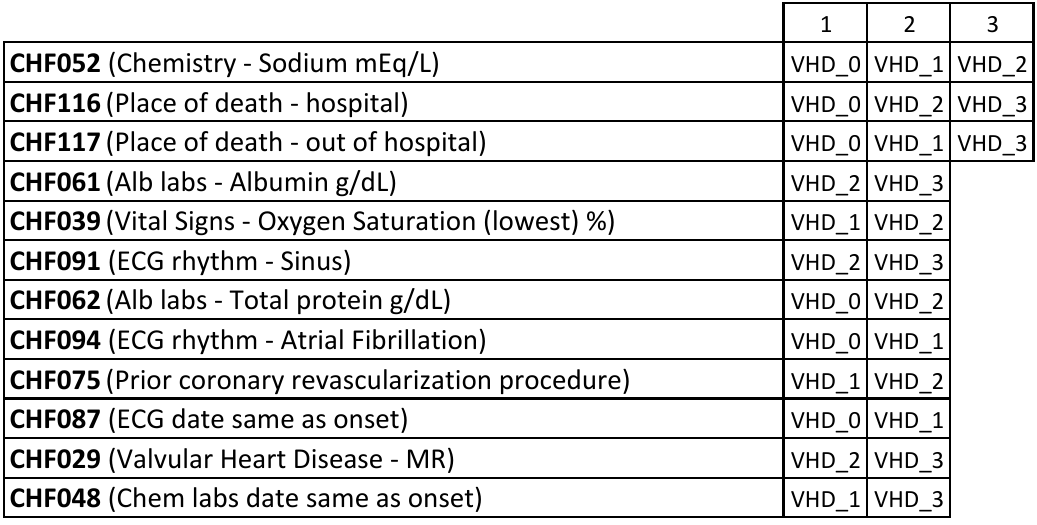}%
  \label{fig:evaluation:revenue}%
}\qquad
\subfloat[RFs Specific to One Subgroup, i.e., subpopulation-level RFs]{%
  \includegraphics[width=0.4\columnwidth]{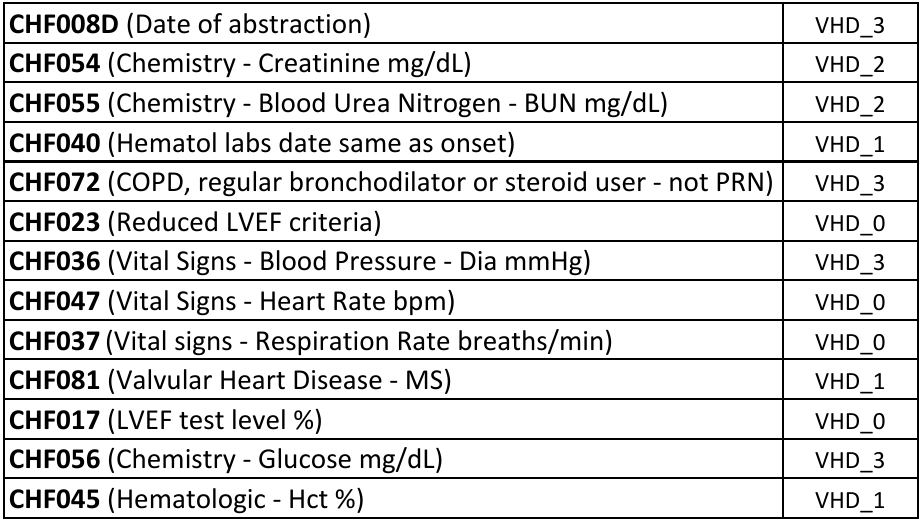}%
 \label{fig:evaluation:avgPrice}%
}

\caption{Cognitive health risk factor analysis result using the MTL model for the \textbf{\textit{FHS}} dataset. The top 10 risk factors (RFs) for cognitive health were selected from each subpopulation (i.e., the participants with various valvular heart disease level) to summarize the set of tasks that shared the same RF in order to generate this result. Subpopulation-level RFs can be found in the same row representing participants with the same level of CHD. For example, CHF045 (i.e., Hematologic - Hct \%), is the subpopulation-specific RF in the VHD level 3 subpopulation shown in the last row. Note that, the ordering of RFs in this figure is based on the number of tasks that share the same RF. However, the ordering of RFs that have the same number of tasks associated with them is arbitrary.
	 \label{fig:statesfhs} }
\end{figure*}

\begin{figure}[ht!]
	\begin{center}
			\includegraphics[width=1\columnwidth]{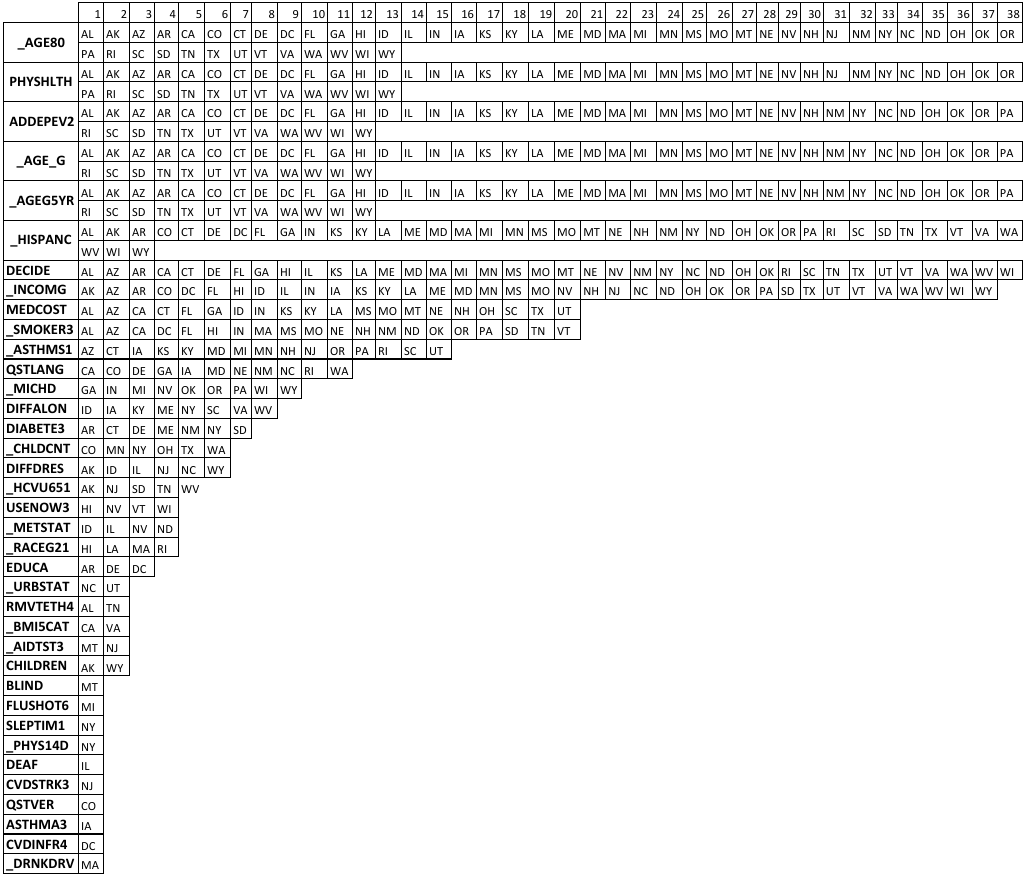}\vspace{-0.3cm}
	\end{center}
	\caption{\small Mental health risk factor analysis result using the CMTL model at subpopulation-level and population-level for \textbf{\textit{BRFSS}} dataset. Top 10 risk factors (RFs) for mental health were selected from each subpopulation (i.e., the participants living in each state/district). Geographic information is represented by abbreviations of states/districts in various colors. Subpopulation-level RFs can be found in the same row, where one interested state/district appears. For example, FLUSHOT6, adult flu shot/spray past 12 months, is the state-specific RF for Michigan shown at the last row in this figure. Note that, the RFs with \_ in front of the names indicated they were calculated variables in BRFSS.
	 } \label{fig:br1}
\end{figure}

\paragraph*{Multi-Level Risk Factor Analysis Results}
Figure \ref{fig:statesfhs} shows the results of two-level (i.e., two/three-subpopulations and subpopulation-levels) risk factor analysis using MTL in the FHS dataset. We also presented three-level (i.e., each subpopulation, each subgroup of the population and the whole population) risk factor analysis using clustered multi-task learning (CMTL) with the BRFSS dataset in Figure \ref{fig:br1} and Table \ref{tab:cmtl}. The FHS and BRFSS datasets employed a comprehensive set of RFs which represented seven different categories, as shown in Table \ref{ct1}. 


In Figure \ref{fig:br1}, population-levels RFs that include all, or almost all, of the US states are shown at the top. Further down the figure RFs are shown that cover fewer states with RFs that are significantly related to only one state (e.g., ASTHMA 3 and Iowa) being shown at the bottom of the figure. 

\begin{figure}[ht!]
	\begin{center}
			\includegraphics[width=0.5\columnwidth]{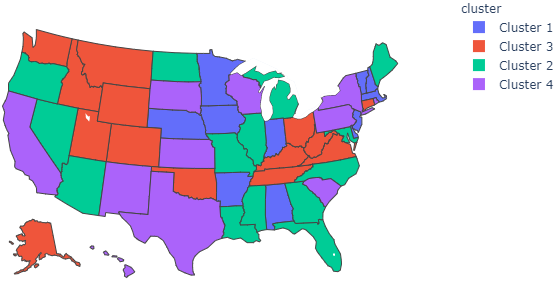}
	\end{center}
	\caption{ Clustering result of four clusters using the CMTL model. Note that, different color represents different cluster for \textbf{\textit{BRFSS}} dataset.
	 \label{fig:r4} }
\end{figure}

We conducted three-level risk factor analysis using the CMTL model on the BRFSS dataset. The most detailed level involved 51 tasks (i.e., each state as one task) and the 51 models were trained simultaneously. As the number of tasks is not small compared with only four tasks in FHS, CMTL was implemented for the clustering of tasks and risk factor analysis synchronously. Since there are four census regions\footnote{\url{https://web.archive.org/web/20130921053705/http://www.census.gov/geo/maps-data/maps/pdfs/reference/us_regdiv.pdf}} in the U.S., i.e., Northeast, Midwest, West and South, we set four clusters for clustering in the CMTL model and the result is shown in Figure \ref{fig:r4}. 

We presented the results of  \textbf{\textit{three-level risk factor analysis (i.e., country-level, four-census-regions-level and state-level)}} for mental health using the CMTL model on the BRFSS dataset in different formats: 1) Results at the cluster-of-subpopulations-level (four-census-regions-level) are shown in Table \ref{tab:cmtl}. 2) Results at the level of individual states (lower part of the figure), and U.S.-wide (upper part of the figure) are shown in Figure \ref{fig:br1}. In Table \ref{tab:cmtl}, the first column represented the cluster numbers that can be referred to Figure \ref{fig:r4} with the other column showing the names of RFs.

\begin{table}[ht!]
\centering
\caption{Categories of risk factors with examples.}
\label{ct1}
\resizebox{\textwidth}{!}{%
\begin{tabular}{|l|l|l|}
\hline
  & Category         & Examples                                                     \\ \hline
1 & Health (both physical and mental) & BMI, blood pressure, blood glucose, mental health status                                   \\ \hline
2 & Lifestyle        & Sleep, smoking, drinking, walking                            \\ \hline
3 & Functional       & Memory, concentrating, errands lone                          \\ \hline
4 & Diagnosis        & Having cancer, stroke, diabetes, asthma, chronic disease     \\ \hline
5 & Social behaviors & Phone usage, driving, seatbelt usage \\ \hline
6 & Demographic characteristics       & Age, family size, educational levels, income, employment, races, health insurance coverage \\ \hline
7 & Other            & Place of death, medical cost, dates, death status            \\ \hline
\end{tabular}%
}
\end{table}
\begin{table}[ht!]
\centering
\caption{Top 10 selected RFs and their corresponding category numbers from our proposed MTL model and six linear regression based STL methods (please refer to Section Discussion for the details of corresponding categories of RFs in Table \ref{ct1}) for \textbf{\textit{FHS}} dataset. Note that, category numbers are shown within parenthesis besides the RFs and their descriptions can be referred to Table \ref{ct1}.\vspace{0.2cm}}
\label{tab:stlfhs}
\resizebox{\textwidth}{!}{%
\begin{tabular}{|l|l|}
\hline
Method Name & Top 10 RFs                                                                   \\ \hline
MTL         & CHF052 (1), CHF116 (7), CHF117 (7), CHF061 (1), CHF039 (1), CHF091 (1), CHF062 (1), CHF094 (1), CHF075 (1)  \\ \hline
STL         & MAXCOG (1), EXAM (7), EFLT50, CHF117 (7), CHF116 (7), CHF115 (7), CHF114 (7), CHF113 (1), CHF112 (1), CHF111 (1) \\ \hline
\end{tabular}%
}
\end{table}

\begin{table}[ht!]
\centering
\caption{Top 10 selected RFs and their corresponding category numbers from our proposed CMTL model and six STL methods (please refer to Section Discussion for the details of corresponding categories of RFs in Table \ref{ct1}) for \textbf{\textit{BRFSS}} dataset. }
\label{tab:stl}
\resizebox{\textwidth}{!}{%
\begin{tabular}{|l|l|}
\hline
Method Name & Top 10 RFs                                                                                                                               \\ \hline
CMTL        & PHYSHLTH (1), \_AGE80 (6), ADDEPEV2 (4), \_AGEG5YR (6), \_HISPANC (6), DECIDE (3), \_INCOMG (6), \_SMOKERS (2) , MEDCOST (7)             \\ \hline
STL         & CHCKDNY1 (4), \_TOTINDA (2), \_AGE\_G (6), \_AGE65YR (6), \_PHYS14D (1),  \_MENT14D (1), DIFFWALK (2), HLTHPLN1 (1), INCOME (6), CVD (4) \\ \hline
\end{tabular}%
}
\end{table}

\begin{table}[ht!]
\centering
\caption{Mental health risk factor analysis result at the four-census-regions-level (subgroup-of-population-level) using the CMTL model for \textbf{\textit{BRFSS}} dataset, i.e., RFs shared by each cluster of subpopulations. Note that, cluster 1, 2, 3, and 4 are equal to the clusters with colors blue, green, red and purple in Figure \ref{fig:r4}, respectively. }
\label{tab:cmtl}
\resizebox{\textwidth}{!}{%
\begin{tabular}{|l|l|}
\hline
Clutser Number & Subgroup-of-population/Cluster-of-subpopulation level RFs \\ \hline
1, 2, 3, 4 & PHYSHLTH, \_AGE80, \_AGE\_G, ADDEPEV2, \_AGEG5YR, \_HISPANC, DECIDE, \_INCOMG, \_SMOKER3, MEDCOST, \_ASTHMS1, QSTLANG,  \_MICHED, DIFFALON, DIABETES \\ \hline
1, 2, 3        & DIFFDRES                           \\ \hline
1, 2, 4        & USENOW3, \_RACEG21                 \\ \hline
1, 3, 4        & \_CHLDCNT, \_HCVU651               \\ \hline
1, 2           & EDUCA                              \\ \hline
1, 3           & RMVTETH4, \_AIDTST3                \\ \hline
2, 3           & \_METSTAT, \_URBSTAT               \\ \hline
3, 4           & \_BMI5CAT                          \\ \hline
1              & \_DRNKDRV, ASTHMA3, CVDSTRK3       \\ \hline
2              & DEAF, FLUSHOT6                     \\ \hline
3              & CHILDREN, BLIND, QSTVER, CVDINFR4  \\ \hline
4              & SLEPTIM1, \_PHYS14D                \\ \hline
\end{tabular}%
}
\end{table}
Since STL trains model independently, it did not seem reasonable to train and combine 51 independent models and then summarize these independent subpopulation-level RFs results to obtain a population-level ranked list of RFs. Thus, we only trained a population-level model using each STL method and compared it with the results obtained by the MTL methods in  Table \ref{tab:stlfhs} and Table \ref{tab:stl} for the FHS and BRFSS datasets, respectively. We obtained the population-level RFs from MD-MTL based on the number of tasks that shared the same RF, while the RFs from STL methods were ranked based on the weight of each variable. For example in Table \ref{tab:stl}, the population-level RFs from the CMTL method were ranked based on the number of U.S. states/district that shared the same RF. 


Since it was not obvious how to efficiently combine information concerning 60 RFs across the STLs, we used a voting strategy to extract the top 10 RFs from the STL methods shown in Table \ref{tab:stlfhs} and Table \ref{tab:stl}. More details are provided in Figure \ref{fhs2} and Figure \ref{brf2} in the Appendices\footnote{\url{https://github.com/wanglu61/MD-MTL-supportive-materials}}, where we picked only one RF at each ranking from the six linear regression based STL models. The ranking method is illustrated with the following example. As shown in Figure \ref{fhs2} in the Appendices, at the first ranking order (i.e., for the first row of the tabulation), Lasso, LassoLars and Elastics all recognized MAXCOG as the top 1 RF, so we picked MAXCOG as the top 1 RF for STL methods in Table \ref{tab:stlfhs}.


\section*{Discussion}
\label{dis}
\paragraph{Disease scores prediction results discussion} In both the FHS and BRFSS datasets, MTL generally outperformed the STL models. The results shown in Table \ref{fhs1} and Table \ref{brfss1} indicated that for every task MTL either yielded the best outcome or was within 0.1 MAE of the best outcome, showing essentially equivalent performance to the best performing STL in the worst case and outperforming all the STL methods otherwise. MTL also outperformed all the STL methods for evaluating the total data set in both cases (FHS and BRFSS).

\vspace{-0.5cm}
\paragraph{Risk factor analysis results discussion} MTL and CMTL performed multi-level risk factor analysis and identified more risk factors (RFs) at the level of individual states, e.g.,\_PHYS14D (i.e., computed physical health status) is an RF unique to New York state that was identified by CMTL but not STL. The results of risk factor analysis for psychological status using both MTL and CMTL on the FHS and BRFSS datasets also showed that multiple tasks were related in terms of shared RFs. In Figure \ref{fig:statesfhs}, no RF was shared by all tasks, demonstrating that there is no population-level RF in the FHS dataset.

The results in Figure \ref{fig:br1} and Table \ref{tab:cmtl} covered three levels of risk factor analysis, i.e., state or subpopulation, U.S.-country-wide or population, and four-census-regions (i.e., subgroup-of-population-level). We can see that both Figure \ref{fig:br1} and Table \ref{tab:cmtl} provide the list of RFs at the population-level from summarizing the top 10 RFs from each state and each cluster of states. Figure \ref{fig:br1} and Table \ref{tab:cmtl} showed five and 15 population-level RFs, respectively. It is because after the clustering of states, the subgroup of population was bigger than subpopulation. The three-level risk factor analysis of the BRFSS data demonstrated that RFs for mental health from different perspectives reflected different geographic groupings. 

As psychological status is a multi-faceted outcome, the RFs are also diverse. In Table \ref{tab:stl}, we can see that CMTL identified RFs for six out of seven categories whereas STL methods identified RFs for only four of the categories. Please refer to the codebooks for the detailed description of RFs in FHS\footnote{\url{https://github.com/wanglu61/MD-MTL-supportive-materials/blob/master/FHS_Data\%20Dictionary.pdf}} and BRFSS\footnote{\url{https://www.cdc.gov/brfss/annual_data/2018/pdf/codebook18_llcp-v2-508.pdf}}. 

Please also refer to the online Appendices at \url{https://github.com/wanglu61/MD-MTL-supportive-materials} for more information.

\section*{Conclusion}
\label{conc}
Although health data is well studied using STL based approaches, STL fails to model the heterogeneity of subpopulations, and the grouping structure of the population. To overcome these limitations, we developed an ensemble machine learning package to conduct disease scores prediction and multi-level risk factor analysis in multiple subgroups of patients within the MTL framework. Our experimental results on one small semi-public and one large public dataset, i.e., the FHS and BRFSS datasets, show that MTL models outperform STL models in terms of prediction performance and multi-level risk factor analysis. In summary, MD-MTL can be used to improve prediction in datasets where there are meaningful subgroups of cases. We continue to develop MD-MTL and intend it to be an online health data preprocessing and analytics tool. In order to achieve this goal, we plan to add an improved user interface, new data preprocessing techniques, and data visualization.
\section*{Acknowledgement}
Funding for this research was provided by an NSERC Discovery Grant ("Dynamic Control of Task Demands in Mobile Contexts using Sensor Data and Adaptive User Models") to the third author. 

\makeatletter
\renewcommand{\@biblabel}[1]{\hfill #1.}
\makeatother

\bibliographystyle{unsrt}

\newpage
\appendixpage
\section{Outcome Variable Distribution in FHS and BRFSS datasets}
We presented the outcome variable distribution for both FHS and BRFSS datasets:
\begin{figure}[htp]
	\begin{center}
			\includegraphics[width=0.5\columnwidth]{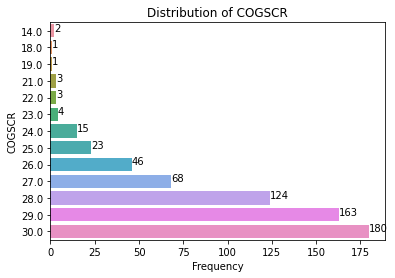}
	\end{center}
	\caption{\small \textbf{The outcome variable COGSCR distribution for FHS dataset.}
	 \label{fhs3} }
\end{figure}

\begin{figure}[ht!]
	\begin{center}
			\includegraphics[width=0.6\columnwidth]{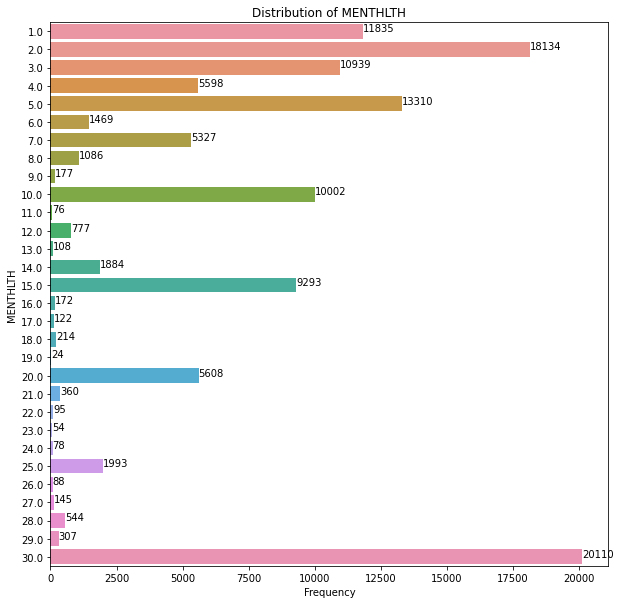}
	\end{center}
	\caption{\small \textbf{The outcome variable MENTHLH for BRFSS dataset.}
	 \label{brf3} }
\end{figure}

\section{Optimization in MTL Algorithm}
Fast iterative shrinkage thresholding algorithm (FISTA) shown in Algorithm \ref{algo1} is implemented to optimize the $l_{2,1}$-norm regularization problem in Eq.(\ref{eq4}) with the general updating steps:
\begin{equation}
\label{o3}
\Phi^{(l+1)}=\pi_P(S^{(l)}-\frac{1}{\gamma^{(l)}}\mathcal{L}'  (S^{(l)})),
\end{equation}
where $l$ is the iteration index, $\frac{1}{\gamma^{(l)}}$ is the possible largest step-size that is chosen by line search and $\mathcal{L}'  (S^{(l)})$ is the gradient of $\mathcal{L} (\cdot)$ at search point $S^{(l)}$. $S^{(l)}=\Phi^{(l)}+\alpha^{(l)}(\Phi^{(l)}-\Phi^{(l-1)})$ are the search points for each task, where $\alpha^{(l)}$ is the combination scalar. $\pi_P(\cdot)$ is $l_{2,1}-$regularized Euclidean projection shown as:
\begin{equation}
\label{o4}
\pi_P (H(S^{(l)}))=\min_{\Phi} \frac{1}{2}||\Phi- H(S^{(l)})||_F^2+\lambda ||\Phi||_{2,1},
\end{equation}
where $H(S^{(l)})= S^{(l)}-\frac{1}{\gamma^{(l)}}\mathcal{L}' (S^{(l)})$ is the gradient step of $S^{(l)}$.
A sufficient scheme that solves Eq.(\ref{o4}) has been proposed as Theorem \ref{the:1}.
\begin{theorem}
\label{the:1}
$\hat{\Phi}$'s primal optimal point in Eq.(\ref{o4}) can be calculated with $\lambda$ as:
\begin{align}
\label{eq:sp}
\small
\hat{\Phi}_j\!=\!\!\left\{\begin{array}{rcl}
\!\!\!\!\!\left(1\!-\frac{\lambda}{\parallel H(S^{(l)})_j \parallel_2}\right)\!H(S^{(l)})_j &\! \!\mbox{if}\!\!\!& \lambda>0,\parallel H(S^{(l)})_j \parallel_2>\lambda\\
0 &\! \!\mbox{if}\!\! \!& \lambda>0,\parallel H(S^{(l)})_j \parallel_2 \leq \lambda\\
H(S^{(l)})_j &\!\! \mbox{if}\!\! \!& \lambda=0,
\end{array}\right.
\end{align}
where $H(S^{(l)})_j$ is the $j^{th}$ row of $H(S^{(l)})$ and $\hat{\Phi}_j$ is the $j^{th}$ row of $\hat{\Phi}$.
\end{theorem}

\begin{algorithm}[htbp]
\SetAlgoLined
\SetKw{Initialize}{Initialize}
\KwIn{Input variables $\{X_1,X_2,\cdots, X_T\}$, output variable $Y$ across all $T$ tasks, initialization of feature weights $\Phi^{(0)}$ and $\lambda$}
\KwOut{$\hat{\Phi}$}
\BlankLine
\Initialize: $\Phi^{(1)}=\Phi^{(0)}$, $d_{-1}=0$, $d_0=1$,$\gamma^{(0)}=1$,$l=1$\;
\Repeat{Convergence of $\Phi^{(l)}$}{
Set $\alpha^{(l)}=\frac{d_{l-2}-1}{d_{l-1}}$, $S^{(l)}=\Phi^{(l)}+\alpha^{(l)}(\Phi^{(l)}-\Phi^{(l-1)})$\;
\For{$j=1,2,\cdots J $}{
Set $\gamma=2^j\gamma_{l-1}$\;
Compute $\Phi^{(l+1)}=\pi_P(S^{(l)}-\frac{1}{\gamma^{(l)}}\mathcal{L}'  (S^{(l)}))$\;
Compute $Q_{\gamma}(S^{(l)},\Phi^{(l+1)})$\;
\If{$\mathcal{L}  (\Phi^{(l+1)}) \leq Q_{\gamma}(S^{(l)},\Phi^{(l+1)})$}{
$\gamma^{(l)}=\gamma$, \textbf{break} \;
}
}
$d_l=\frac{1+\sqrt{1+4d^2_{l-1}}}{2}$\;
$l=l+1$\; 
}
$\hat{\Phi}=\Phi^{(l)}$\;
\caption{Fast iterative shrinkage thresholding algorithm (FISTA) for optimizing the $l_{2,1}$-norm regularization problem.}
\label{algo1}
\end{algorithm}
\DecMargin{1.5em}

From the $4^{th}$ line to the $11^{th}$ line in Algorithm \ref{algo1}, the optimal $\gamma^{(l)}$ is chosen by the backtracking rule. And $\gamma^{(l)}\geq b$, where $b$ is the Lipschitz constant of $\mathcal{L} ( \cdot)$ at search point $S^{(l)}$, which means $\gamma^{(l)}$ is satisfied for $S^{(l)}$ and $\frac{1}{\gamma^{(l)}}$ is the possible largest step size. 

At the $7^{th}$ line in Algorithm \ref{algo1}, tangential line of $\mathcal{L}  (\cdot)$ at search point $S^{(l)}$, denoted as $Q_{\gamma}(S^{(l)},\Phi^{(l+1)})$, is computed by:
\begin{align}
Q_{\gamma}(S^{(l)},\Phi^{(l+1)})&=\mathcal{L}  (S^{(l)})+\frac{\gamma}{2}\parallel \Phi^{(l+1)}-S^{(l)}\parallel ^2\nonumber\\
&+\langle \Phi^{(l+1)}-S^{(l)}, \mathcal{L}' (S^{(l)})  \rangle .\nonumber
\end{align}
\section{Optimization in CMTL Algorithm}
In Eq.(\ref{eq:cCMTL}), the equation is conjointly convex with respect to (w.r.t.) $C$ and $\Phi$, which is an convex unconstrained smooth optimization problem w.r.t. $C$. We iteratively update the gradient step of the aforementioned optimization problem in order to find the global optimum w.r.t. $C$: 
\begin{equation}
G_\Phi=S-\frac{1}{\gamma} [\bigtriangledown \mathcal{L}(S_\Phi)+ 2\rho_1\eta(1+\eta)(\eta I+C_S)^{-1}S^T],
\end{equation}
where $S_\Phi$ is the search point of $\Phi$ that is defined as $S_\Phi^{(l)}=\Phi^{(l)}+\alpha^{(l)}(\Phi^{(l)}-\Phi^{(l-1)})$. The search point of $C$ is denoted as $C_S$, which can be similarly updated as $C_S^{(l)}=C^{(l)}+\alpha^{(l)}(C^{(l)}-C^{(l-1)})$ at the $l^{th}$ iteration. $\bigtriangledown\mathcal{L}(S)$ is the gradient of $\mathcal{L}(S)$ that is calculated as:
\begin{equation}
\bigtriangledown \mathcal{L}(S)=\left[\frac{l^{'}(S_1)}{N_1},\frac{l^{'}(S_2)}{N_2},\cdots,\frac{l^{'}(S_T)}{N_T}\right].
\end{equation}

Similarly, in the optimization of MTFL, FISTA is also implemented for optimizing the CMTL, except the line 6 is replaced with the corresponding proximal operator that is solved by the following steps. To optimize the convex set $C$, we need to solve a convex constrained minimization problem, which is formulated with its corresponding proximal operator and calculated using its gradient step, denoted as $G_C$, at the search point $C_S$:
\begin{equation}
\label{eq:proximalC}
\min_{C}\parallel \!\!C - G_C \!\!\parallel_F^2,\quad\! \text{s.t.} \quad\! \text{tr}(C)=K, C \preceq I, C\in \mathbb{S}^T_{+}. 
\end{equation}
We can compute the $G_C$ by:
\begin{equation}
\label{eq: GW}
G_C=C_S+\frac{\rho_1\eta(1+\eta)}{\gamma} S^TS(\eta I+C_S)^{-2}.
\end{equation}
 A solution of Eq.(\ref{eq:proximalC}) is proposed and summarized in the following theorem.  

\begin{theorem}
\label{the:cCMTL}
Let $G_T=V \hat{\Sigma} V^T$ be the eigen-decomposition of gradient step $G_C \in \mathbb{S}^{T \times T}$, where $\hat{\Sigma} =\text{diag}(\hat{\sigma}_1,\cdots, \hat{\sigma}_T) \in \mathbb{R}^{T \times T}$ and $V\in \mathbb{R}^{T \times T}$ is orthonormal. The optimization problem is formulated as:
\begin{align}
& \underset{\{\sigma_m\}}{\text{min}} &\!\!\!\! &\sum_{t=1}^T (\sigma_t-\hat{\sigma}_t)^2. \nonumber\\
& \text{s. t.} &\!\!\!\! \!\! &  \sum_{t=1}^T\sigma_t=K, \;0\leq\sigma_t\leq 1,\; \forall t=1,\cdots,T 
\end{align}
Let $\Sigma^{*}=diag(\sigma_1^{*},\cdots, \sigma_T^{*}) \in \mathbb{R}^{T \times T}$, so that the optimal solution of the above optimization problem is $\{\sigma_1^{*},\cdots, \sigma_T^{*} \}$.
As a result, the proximal operator's optimal solution in Eq.(\ref{eq:proximalC}) is calculated as $\hat{T}=V \Sigma^{*}V^T$.
\end{theorem} 

\section{Risk Factor Analysis - Additional Results}
We presented the additional results of risk factor analysis for FHS and BRFSS datasets:
\begin{figure}[htp]
	\begin{center}
			\includegraphics[width=1\columnwidth]{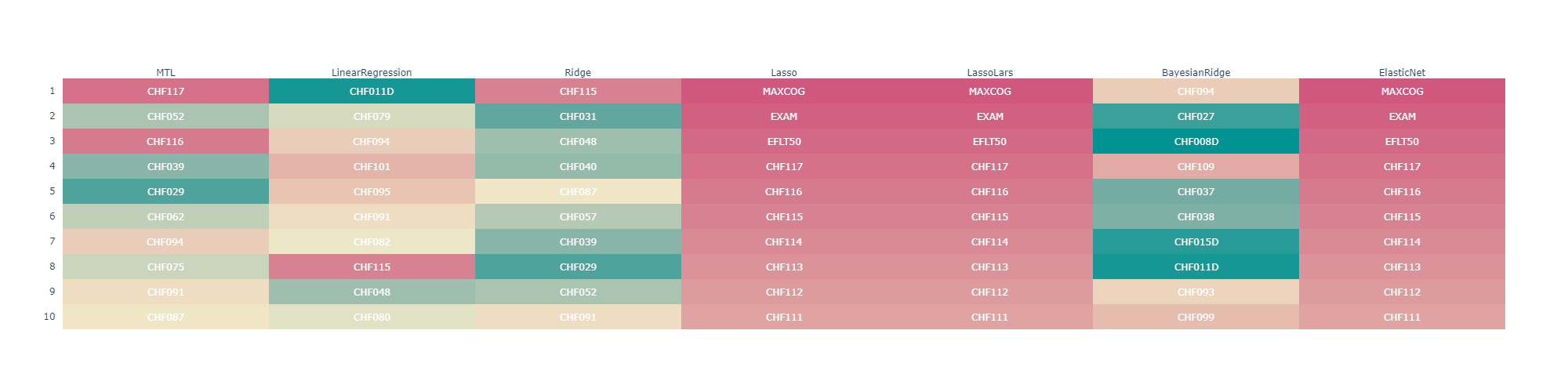}
	\end{center}
	\caption{\small \textbf{Top 10 selected RFs and their corresponding category numbers from our proposed MTL method and seven STL methods for FHS dataset. Please zoom in for clear visualization. }
	 \label{fhs2} }
\end{figure}

\begin{figure}[ht!]
	\begin{center}
			\includegraphics[width=1\columnwidth]{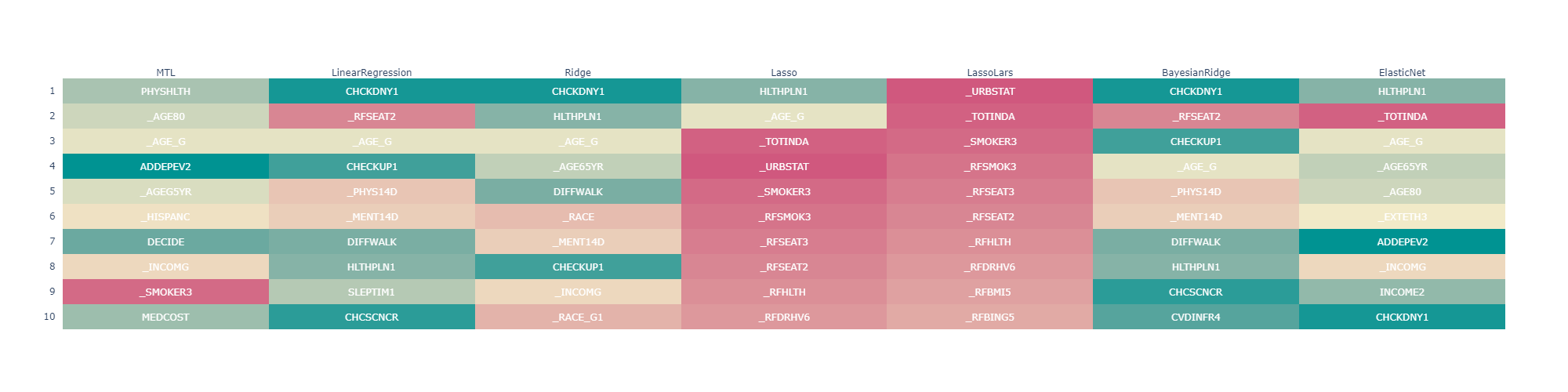}
	\end{center}
	\caption{\small \textbf{Top 10 selected RFs and their corresponding category numbers from our proposed CMTL method and six STL methods for BRFSS dataset. Please zoom in for clear visualization. }
	 \label{brf2} }
\end{figure}

\end{document}